\documentclass{article} % For LaTeX2e
\usepackage{iclr2016_conference,times}
\usepackage{hyperref}
\usepackage{url}
\usepackage{amsmath,graphicx}
\usepackage[ruled,vlined]{algorithm2e}
\usepackage{amsfonts}
\usepackage{amssymb}

\title{Deep multi-scale video prediction beyond mean square error}
\author{Michael Mathieu\textsuperscript{1, 2}, Camille Couprie\textsuperscript{2} \&
Yann LeCun\textsuperscript{1, 2}\\
\textsuperscript{1}New York University\\
\textsuperscript{2}Facebook Artificial Intelligence Research\\
\texttt{mathieu@cs.nyu.edu, \{coupriec,yann\}@fb.com}\\
}

\newcommand{\NLL}{L_{bce}}
\newcommand{\omitme}[1]{}

\newcommand{\nscales}{N_{\mathrm{scales}}}

\iclrfinalcopy % Uncomment for camera-ready version

\begin{document}

\maketitle

\begin{abstract}
Learning to predict future images from a video sequence involves the
construction of an internal representation that models the image evolution
accurately, and therefore, to some degree, its content and dynamics.
This is why
pixel-space video prediction may be viewed as a promising avenue for unsupervised feature learning.
In addition, while optical flow has been a very studied problem in computer vision for a long
time, future frame prediction is rarely approached. Still, many vision applications could benefit from the knowledge
of the next frames of videos, that does not require the complexity of tracking
every pixel trajectory.
In this work, we train a convolutional network to generate future frames given
an input sequence. To deal with the inherently blurry
predictions obtained from the standard Mean Squared Error (MSE) loss function, we propose
three different and complementary feature learning strategies: a multi-scale
architecture, an adversarial training method, and an image gradient difference loss function.
We compare our predictions to different published results based on recurrent neural
networks on the UCF101 dataset.
%We
%benchmark these different predictions approaches using image quality measures
%such as the peak signal to noise ratio and an image sharpness index.

\end{abstract}
%In this work, we focus on pixel-space video prediction as a proxy
%for unsupervised feature learning.

%***************************************
\section{Introduction}
%***************************************

Unsupervised feature learning of video representations is a promising direction
of research because the resources are quasi-unlimited and the progress remaining
to achieve in this area are quite important.
In this paper, we address the problem of frame \emph{prediction}. A significant difference
with the more classical problem of image reconstruction \citep{vincent2008extracting,le2013building}
is that the ability of a model to predict future frames requires to build accurate,
non trivial internal representations, even in the absence of other constraints (such as sparsity).
Therefore, we postulate that the better the predictions of such system are,
the better the feature representation should be.
Indeed, the work of \citet{SrivastavaMS15} demonstrates
that learning representations by predicting the next sequence of image features
improves classification results on two action recognition datasets.
In this work, however, we focus on predicting directly in pixel space and try to
address the inherent problems related to this approach.

Top performing algorithms for action recognition exploit the temporal
information in a supervised way, such as the 3D convolutional network of
\citet{Tran2015C3D}, or the spatio-temporal convolutional model of
\citet{Simonyan2014flowAction}, which can require months of training, and heavily labeled datasets. This could be reduced using unsupervised
learning. The authors in \citep{Wang2015UnsupPatchVideo} compete with supervised learning
performance on ImageNet, by using a siamese architecture
\citep{bromley1993siamese} to mine positive and negative examples from patch
triplets of videos in an unsupervised fashion.
Unsupervised learning from video is also exploited in the work of
\citet{Vondrick2015Future}, where a convolutional model is trained to predict
sets of future possible actions, or in \citep{Jayaraman2015egomotion} which
focuses on learning a feature space equivariant to ego-motion. \citet{GoroshinML15} trained a convolutional network to learn to linearize
motion in the code space and tested it on the NORB dataset. Beside unsupervised learning, a video predictive system may
find applications in robotics \citep{kosaka1992fast}, video compression
\citep{ascenso2005improving} and inpainting \citep{Flynn2015deepstereo}
to name a few. %\textcolor{red}{[TODO check if ref covers it all: no, we have refs for robotic
  %and compression]}

Recently, predicting future video sequences appeared in different settings:
\citet{Ranzato2014videolanguage} defined a recurrent network architecture
inspired from language modeling, predicting the frames in a discrete space
of patch clusters.
\citet{SrivastavaMS15} adapted a LSTM model \citep{Hochreiter1997LSTM} to future frame prediction.
\citet{Oh2015ActionPredictionAtari} defined an action conditional auto-encoder
model to predict next frames of Atari-like games.
In the two works dealing with natural images, a blur effect is observed in the
predictions, due to different causes. In \citep{Ranzato2014videolanguage}, the
transformation back and forth between pixel and clustered spaces involves the averaging
of 64 predictions of overlapping tilings of the image, in order to avoid a blockiness effect
in the result. Short term results from \citet{SrivastavaMS15}
are less blurry, however the $\ell_2$ loss function inherently
produces blurry results.
Indeed, using the $\ell_2$ loss comes from the assumption that the data is drawn from
a Gaussian distribution, and works poorly with multimodal distributions.

%Latent variables used for 3D rendering in \cite{Kulkarni2015latentGraphics}

%Next frame generation : blurry predictions.

In this work, we address the problem of lack of sharpness in the predictions.
We assess different loss functions, show that generative adversarial training
\citep{Goodfellow2014adversarial,Denton2015deep} may be successfully employed for
next frame prediction, and finally introduce a new loss based on the image gradients,
designed to preserve the sharpness of the frames.
Combining these two losses produces the most visually satisfying results.

%Related work in terms of regularization of convnet feature visualization \cite{MahendranV14,SimonyanVZ13}.

%In \cite{Ranzato2014videolanguage}, the blur is partly limited by operating in a
%$k$-mean embeding of images patches.

%Predictions in the feature space \cite{Vondrick2015anticipating}

Our paper is organised as follows: the model section describes the different
model architectures: simple,
multi-scale, adversarial, and presents our gradient difference loss function.
The experimental section compares the proposed architectures and losses on
video sequences from the Sports1m dataset of \citet{KarpathyCVPR14}
and UCF101 \citep{Soomro2012UCF101}.
We further compare our results with \citep{SrivastavaMS15} and
\citep{Ranzato2014videolanguage}. We measure the quality of image generation
by computing similarity and sharpness measures.

 %See \citet{Hinton06} for more information
% Deep learning shows promise to make progress towards AI~\citep{Bengio+chapter2007}
%**********************************************************************************
\section{\label{section:model}Models}
%**********************************************************************************

Let $Y=\{Y^1,...,Y^n\}$ be a sequence of frames to predict from input frames
$X=\{X^1,...,X^m\}$ in a video sequence. Our approach is based on a convolutional network
\citep{lecun1998gradient}, alternating convolutions
and Rectified Linear Units (ReLU) \citep{nair2010rectified}.

\begin{figure}[htb]
  \caption{A basic next frame prediction convnet}
  \begin{center}
    \begin{tabular}{lcccccr}
      Input & First & Second& Third &Fourth& Fifth &Output\\
      $X$ &  feature map & feature map & feature map & feature map & feature
map & $G(X)$
\end{tabular}\\
    \includegraphics[width=1\textwidth]{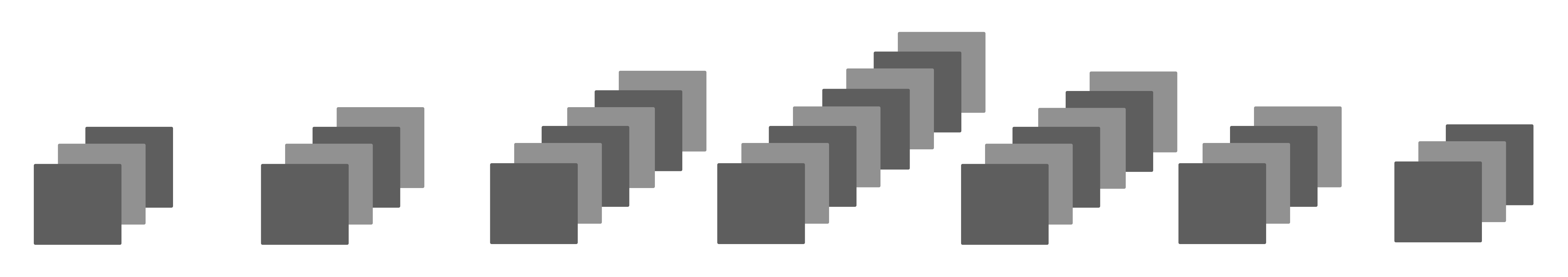} \\
   conv. ~ ReLU ~ conv. ~ ReLU  ~ conv. ~ ReLU ~
    conv. ~ ReLU ~ conv. ~Tanh\\
    \end{center}
\label{fig:convnet}
\end{figure}

Such a network $G$, displayed in Figure~\ref{fig:convnet}, can be trained to predict
one or several concatenated frames $Y$ from the concatenated
frames $X$ by minimizing a distance, for instance $\ell_p$ with $p=1$ or $p=2$,
between the predicted frame and the true frame:
\begin{equation}
  \mathcal{L}_p(X,Y) = \ell_p(G(X), Y) = \|G(X) - Y\|_p^p,
  \label{eq:loss}
\end{equation}
However, such a network has at least two major flaws:\\
%\begin{enumerate}
1. Convolutions only account for short-range dependencies, limited by the size of their
  kernels. However, using pooling
  would only be part of the solution since the output has to be of the same resolution as
  the input. There are a number of ways to avoid the loss of resolution brought about by
  pooling/subsampling while preserving long-range dependencies.
  The simplest and oldest one is to have no pooling/subsampling but many convolution
  layers \citep{jain2007nopooling}.
  Another method is to use connections that ``skip'' the pooling/unpooling pairs, to preserve
  the high frequency information
  \citep{long2014fully,Dosovitskiy2014chairs,Ronneberger2015}.
  Finally, we can combine multiple scales linearly as in the reconstruction process of
  a Laplacian pyramid \citep{Denton2015deep}. This is the approach we use in this paper.\\
2. Using an $\ell_2$ loss, and to a lesser extent $\ell_1$, produces blurry predictions,
  increasingly worse when predicting further in the future. If the probability distribution
  for an output pixel has two equally likely modes $v_1$ and $v_2$, the value $v_{avg}=(v_1+v_2)/2$
  minimizes the $\ell_2$ loss over the data, even if $v_{avg}$ has very low probability.
  In the case of an $\ell_1$ norm, this effect diminishes, but do not disappear,
  as the output value would be the median of the set of equally likely values.
%\end{enumerate}

\subsection{Multi-scale network}

\begin{figure}[htb]
  \caption{Multi-scale architecture}
  \begin{center}
    \includegraphics[width=1\textwidth]{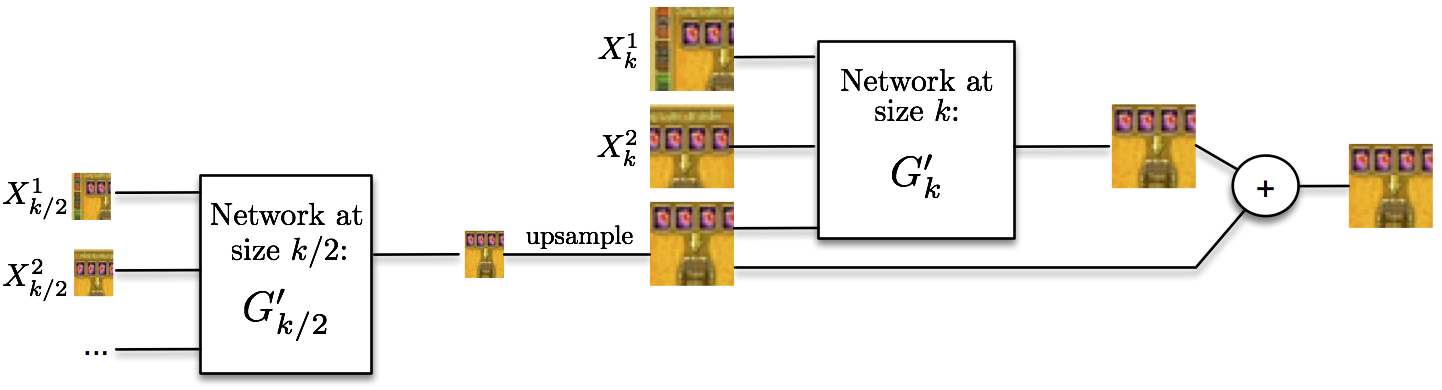}%{mutliscale_next_frame.pdf} \\
    \end{center}
\label{fig:multiscale}
\end{figure}

We tackle Problem 1 by making the model multi-scale.
A multi-scale version of the model is defined as follows:
Let $s_1, \dots, s_{\nscales}$ be the sizes of the inputs of our network. %our network is going to work on.
Typically, in our experiments, we set $s_1=4\times 4$, $s_2=8\times8$,
$s_3=16\times16$ and $s_4=32\times32$.
Let $u_k$ be the upscaling operator toward size $s_k$.
Let $X_k^i$, $Y_k^i$ denote the downscaled versions of $X^i$ and $Y^i$ of size $s_k$,
and $G'_k$ be a network that learns to predict $Y_k-u_k(Y_{k-1})$ from $X_k$ and
a coarse guess of $Y_k$.
We recursively define
the network $G_k$, that makes a prediction $\hat{Y}_k$ of size
$s_k$, by
\begin{equation}
  \hat{Y}_k = G_k(X) =  u_k(\hat{Y}_{k-1}) + G'_k\left(X_{k}, u_k(\hat{Y}_{k-1})\right).
\end{equation}
Therefore, the network makes a series of predictions, starting from the lowest
resolution, and uses the prediction of size $s_k$ as a starting point
to make the prediction of size $s_{k+1}$. At the lowest scale $s_1$,
the network takes only $X_1$ as an input.
This architecture is illustrated on Figure~\ref{fig:multiscale}, and the
specific details are given in Section \ref{section:experiments}.
The set of trainable parameters is denoted $W_G$ and the minimization
is performed via Stochastic Gradient Descent (SGD).

%We refer to the multi-scale model minimizing an $\ell_1$ (respectively $\ell_2$) norm
%as the $\ell_1$ (respectively $\ell_2$) model in the remainder of the paper.
Despite the multi-scale architecture, the search of $Y$ from $X$ without making any
assumption on the space of possible configurations still leads to blurry predictions, because
of Problem 2. In order to further reduce this
effect, the next two sections introduce an adversarial strategy and the image gradient difference loss.

%********************************
\subsection{Adversarial training}
%********************************

Generative adversarial networks were introduced by
\citet{Goodfellow2014adversarial}, where images patches are generated from random
noise using two networks trained simultaneously. In that work, the authors propose
to use a discriminative network $D$ to estimate the probability that a sample comes
from the dataset instead of being produced by a generative model $G$.
The two models are simultaneously
trained so that $G$ learns to generate frames that are hard to classify by $D$,
while $D$ learns to discriminate the frames generated by $G$. Ideally, when $G$
is trained, it should not be possible for $D$ to perform better than chance.

We adapted this approach for the purpose of frame prediction, which constitutes to our
knowledge the first application of adversarial training to video prediction.
The generative model $G$
is typically the one described in the previous section. The discriminative model
$D$ takes a sequence of frames, and is trained to predict the probability that the last frames
of the sequence are generated by $G$. Note only the last frames are either real of generated
by $G$, the rest of the sequence is always from the dataset.
This allows the discriminative model to make use of temporal information, so that
$G$ learns to produce sequences that are temporally coherent with its input.
Since $G$ is conditioned on the input frames $X$, there is variability in the
input of the generator even in the absence of noise, so noise is not a necessity anymore.
We trained the network with and without adding noise and did not observe any
difference. The results we present are obtained without random noise.

Our main intuition on why to use an adversarial loss is that it can, theoretically, address the
Problem 2 mentioned in Section~\ref{section:model}. Imagine a sequence of frames
$X = (X^1,\dots,X^m)$ for which, in the dataset, the next frames can either be
$Y = (Y^1,\dots,Y^n)$ or $Y'=(Y'^1,\dots,Y'^n)$, with equal probability. As explained before,
training the network with an $\ell_2$ loss will result in predicting the average frames
$Y_{avg}=(Y+Y')/2$. However, the sequence $(X, Y_{avg})$, composed of the frames of $X$
followed by the frames of $Y_{avg}$,
is not a likely sequence, and $D$ can discriminate them easily.
The only sequences the model $D$ will not be able to classify as fake
are $(X, Y)$ and $(X, Y')$.

The discriminative model $D$ is a multi-scale convolutional network with a single scalar output.
The training of the pair ($G$, $D$) consists of two alternated steps, described below. For the
sake of clarity, we assume that we use pure SGD (minibatches of size 1), but there is no difficulty
to generalize the algorithm to minibatches of size M by summing the losses over the samples.
\paragraph{Training $\mathbf{D}$:} Let $(X, Y)$ be a sample from the dataset.
Note that $X$ (respectively $Y$) is a sequence of $m$ (respectively $n$) frames.
We train $D$ to classify the input $(X, Y)$ into class $1$ and the input $(X, G(X))$
into class $0$.
More precisely, for each scale $k$, we perform one SGD iteration of $D_k$ while keeping
the weights of $G$ fixed. It is trained with in the target $1$ for the datapoint $(X_k, Y_k)$,
and the target $0$ for $(X_k, G_k(X_k))$. Therefore, the loss function we use to train $D$ is
\begin{equation}
  \mathcal{L}_{adv}^D(X, Y) = \sum_{k=1}^{\nscales}\NLL(D_k(X_k, Y_k), 1) + \NLL(D_k(X_k, G_k(X)), 0)
\end{equation}
where $\NLL$ is the binary cross-entropy loss, defined as
\begin{equation}
  \NLL(Y, \hat{Y}) = - \sum_i \hat{Y}_i \log\left(Y_i\right) + (1 - \hat{Y}_i) \log\left(1-Y_i\right)
\end{equation}
where $Y_i$ takes its values in $\{0,1\}$ and $\hat{Y_i}$ in $[0,1]$.
\paragraph{Training $\mathbf{G}$:} Let $(X, Y)$ be a \emph{different} data sample.
While keeping the weights of $D$ fixed, we perform one SGD step on $G$ to minimize
the adversarial loss:
\begin{eqnarray}
 % \mathcal{L}_{G}(X, Y) = \sum_{k=1}^{\nscales}\big(\lambda_{D}\NLL(D_k(X_k,
  % G_k(X)), 1) + \lambda_{G}||Y_k - G_k(X)||_2^2\big)
  %\mathcal{L}^G(X, Y) &=& \lambda_{adv} \mathcal{L}_{adv}^G(X, Y) + \lambda_{\ell_p} \mathcal{L}_{\ell_p}(X, Y) \\
  \mathcal{L}_{adv}^G(X, Y) &=& \sum_{k=1}^{\nscales}\NLL(D_k(X_k, G_k(X_k)), 1)
  \label{eq:advloss}
\end{eqnarray}
Minimizing this loss means that the generative model $G$ is making the
discriminative model $D$ as ``confused'' as possible, in the sense that $D$
will not discriminate the prediction correctly.
However, in practice, minimizing this loss alone can lead to instability.
$G$ can always generate samples that ``confuse'' $D$, without being close to $Y$. In turn,
$D$ will learn to discriminate these samples, leading $G$ to generate other
``confusing'' samples, and so on.
To address this problem, we train the generator with a combined loss composed of
the of the adversarial loss and the $\mathcal{L}_p$ loss \omitme{(later in the paper we add the
$\mathcal{L}_{gdl}$ loss)}. The generator $G$ is therefore trained to minimize
$\lambda_{adv}\mathcal{L}_{adv}^G + \lambda_{\ell_p}\mathcal{L}_p$.
There is therefore a tradeoff to adjust, by
the mean of the $\lambda_{adv}$ and $\lambda_{\ell_p}$ parameters, between sharp
predictions due to the adversarial principle, and similarity with the ground truth
brought by the second term.
This process is summarized in Algorithm~\ref{algo:adv},
with minibatches of size $M$.

\begin{algorithm}
  Set the learning rates $\rho_D$ and $\rho_G$, and weights $\lambda_{adv}, \lambda_{\ell_p}$.\\
  \While {not converged} {
    {\bf Update the discriminator $D$:}\\
    Get $M$ data samples $(X, Y) = (X^{(1)}, Y^{(1)}), \dots, (X^{(M)}, Y^{(M)})$\\
    $W_D = W_D - \rho_D \sum_{i=1}^M\frac{\partial \mathcal{L}_{adv}^D(X^{(i)}, Y^{(i)})}{\partial W_D}$\\
    {\bf Update the generator $G$:}\\
    Get $M$ new data samples $(X, Y) = (X^{(1)}, Y^{(1)}), \dots, (X^{(M)}, Y^{(M)})$\\
    $W_G = W_G - \rho_G \sum_{i=1}^M\Big(\lambda_{adv}\frac{\partial \mathcal{L}_{adv}^G(X^{(i)}, Y^{(i)})}{\partial W_G}
    + \lambda_{\ell_p}\frac{\partial \mathcal{L}_{\ell_p}(X^{(i)}, Y^{(i)})}{\partial W_G}\Big)$\\
  }
  \caption{Training adversarial networks for next frame generation}
  \label{algo:adv}
\end{algorithm}

%************************************
\subsection{Image Gradient Difference Loss (GDL)}
%************************************

Another strategy to sharpen the image prediction is to directly penalize
the differences of image gradient predictions in the generative loss function.
We define a new loss function, the Gradient Difference Loss (GDL), that can
be combined with a $\ell_p$ and/or adversarial loss function.
%, by simply adding it to \eqref{eq:loss} or \eqref{eq:advloss}.
The GDL function between the ground truth image $Y$, and the
prediction $G(X)=\hat{Y}$ is given by
\begin{multline}
  \mathcal{L}_{gdl}(X,Y) = L_{gdl}(\hat{Y},Y)= \\
   \sum_{i,j} \big| |Y_{i,j}-Y_{i-1,j}| - |\hat{Y}_{i,j}-\hat{Y}_{i-1,j}| \big|^{\alpha} + \big| |Y_{i,j-1}-Y_{i,j}| - |\hat{Y}_{i,j-1}-\hat{Y}_{i,j}| \big|^{\alpha},
\end{multline}
where $\alpha$ is an integer greater or equal to 1, and $|.|$ denotes the
absolute value function.
To the best of our knowledge, the closest related work to this idea is the work
of \citet{MahendranV14}, using a total variation regularization to generate images
from learned features. Our GDL is fundamentally different: In
\citep{MahendranV14}, the total variation takes only the reconstructed frame in
input, whereas our loss penalises gradient differences between the prediction and
the true output. Second, we chose the simplest possible image gradient by
considering the neighbor pixel intensities differences, rather than adopting a
more sophisticated norm on a larger neighborhood, for the sake of keeping the
training time low.

%************************************
\subsection{Combining losses}
%************************************

In our experiments, we combine the losses previously defined with different
weights. The final loss is:
\begin{equation}
  \mathcal{L}(X, Y) = \lambda_{adv}\mathcal{L}_{adv}^G(X, Y) + \lambda_{\ell_p} \mathcal{L}_p(X, Y) + \lambda_{gdl}\mathcal{L}_{gdl}(X, Y)
\end{equation}

%**********************************************************************************
\section{\label{section:experiments}Experiments}
%**********************************************************************************

We now provide a quantitative evaluation of the quality of our video predictions
on UCF101~\citep{Soomro2012UCF101} and Sports1m~\citep{KarpathyCVPR14} video clips.
We train and compare two configurations: (1) We use $4$ input frames to predict one future frame.
In order to generate further in the future, we apply the model recursively by
using the newly generated frame as an input. (2) We use $8$ input frames to produce
$8$ frames simultaneously. This second configuration represents a significantly
harder problem and is presented in Appendix.

\subsection{Datasets}

\omitme{Currently, the two largest publicly available video datasets are UCF101 and
Sports1m.} We use the Sports1m for the training,
because most of UCF101 frames only have a very small portion of the image
actually moving, while the rest is just a fixed background.
We train our network by randomly selecting temporal sequences of patches of
$32 \times 32$ pixels \omitme{from the Sports1m dataset,}
after making sure they show enough movement
(quantified by the $\ell_2$ difference between the frames).
The data patches are first normalized so that their values are comprised
between -1 and 1.

\omitme{
However, although the Sports1m dataset has labeled sports class, we do not provide
classification results on this dataset. Indeed this datasets is so large than producing
significant results would likely take months of trainings, and this is not the point of
our paper. Instead, we transfer our learned feature on the UCF101 dataset, and
train a classifier on top of our features.
% Camille : A rajouter si on a des resultats
}

\subsection{Network architecture}

\omitme{
\begin{table}
  \caption{Different networks}
  \begin{center}
    \begin{tabular}{c|c|c|c|c|c|c|c|}
      \cline{2-8}
      & $\nscales$ & $\lambda_{\ell_p}$ & $p$ & $\lambda_{adv}^G$ & $\lambda_{adv}^D$ & $\lambda_{gdl}$ & $\alpha$ \\
      \hline
      \multicolumn{1}{|c|}{single sc. $\ell_2$} & 1 & 1 & 2 & 0 &  & 0 &  \\
      \multicolumn{1}{|c|}{$\ell_2$}            & 4 & 1 & 2 & 0 &  & 0 &  \\
      \multicolumn{1}{|c|}{$\ell_1$}            & 4 & 1 & 1 & 0 &  & 0 &  \\
      \multicolumn{1}{|c|}{GDL $\ell_1$}        & 4 & 1 & 1 & 0 &  & 1 & 1 \\
      \multicolumn{1}{|c|}{GDL $\ell_2$}        & 4 & 1 & 2 & 0 &  & 1 & 1 \\
      \multicolumn{1}{|c|}{Adv}                 & 4 & 1 & 2 & 1 & 0.05& 0 &  \\
      \multicolumn{1}{|c|}{Adv+GDL}             & 4 & 1 & 2 & 1 & 0.05& 1 & 2 \\
      \hline
    \end{tabular}
  \end{center}
\end{table}
}

\begin{table}
  \caption{Network architecture (Input: 4 frames -- output: 1 frame)}
  \begin{center}
    {\scriptsize \begin{tabular}{ccccc}
        \hline
        Generative network scales  & $G_1$ & $G_2$&  $G_3$ & $G_4$ \\
        \hline
        Number of feature maps &  128, 256, 128 & 128, 256, 128 & 128, 256, 512, 256, 128 & 128, 256, 512, 256, 128 \\
        Conv. kernel size & 3, 3, 3, 3 & 5, 3, 3, 5 & 5, 3, 3, 3, 5 & 7, 5, 5, 5, 5, 7 \\
        \hline
        \hline
        Adversarial network scales  & $D_1$ & $D_2$&  $D_3$ & $D_4$ \\
        \hline
        Number of feature maps &  64 & 64, 128, 128 & 128, 256, 256 & 128, 256, 512, 128 \\
        Conv. kernel size (no padding) & 3 & 3, 3, 3 & 5, 5, 5 & 7, 7, 5, 5 \\
        Fully connected & 512, 256 & 1024, 512 & 1024, 512 & 1024, 512 \\
        \hline
    \end{tabular}}
    \end{center}
  \label{tbl:network}
\end{table}

We present results for several models. Unless otherwise stated, we employed
mutliscale architectures. Our baseline models are using $\ell_1$ and $\ell_2$
losses. The GDL-$\ell_1$ (respectively GDL-$\ell_2$)
model is using a combination of the GDL with $\alpha=1$ (respectively $\alpha=2$)
and $p=1$ (respectively $p=2$) loss; the relative weights $\lambda_{gdl}$ and $\lambda_{\ell_p}$ are both $1$.
The adversarial (Adv) model uses the adversarial loss, with $p=2$
weighted by $\lambda_{adv} = 0.05$ and $\lambda_{\ell_p} = 1$.
Finally, the Adv+GDL model is a combination or the adversarial loss and the GDL,
with the same parameters as for Adv with $\alpha=1$ and $\lambda_{gdl} = 1$.
%These models are summarized in table~\ref{tbl:network2}.

\paragraph{Generative model training:} The generative model $G$ architecture is presented in
Table~\ref{tbl:network}.
It contains padded convolutions interlaced with ReLU non linearities.
A Hyperbolic tangent (Tanh) is added at the end of the model
to ensure that the output values are between -1 and 1. The learning rate
$\rho_G$ starts at $0.04$ and is reduced over time to $0.005$. The minibatch size is set to $4$,
or $8$ in the case of the adversarial training, to take advantage of
GPU hardware capabilities. We train the network on small patches, and since it is fully
convolutional, we can seamlessly apply it on larger images at test time.

\paragraph{Adversarial training:} The discriminative model $D$,
also presented in Table~\ref{tbl:network},
uses standard non padded convolutions followed by fully connected layers and ReLU non linearities.
For the largest scale $s_4$, a $2\times 2$ pooling is added after the convolutions. \omitme{We do not
include pooling earlier in the network since we really care about sharpness.} The
network is trained by setting the learning rate $\rho_D$ to $0.02$.

%\vspace*{-0.3cm}
\subsection{Quantitative evaluations}

%[TODO: update the adversatial resutls]

To evaluate the quality of the image predictions resulting from the different
tested systems, we compute the Peak Signal to Noise Ratio (PSNR) between the
true frame $Y$ and the prediction $\hat{Y}$:
\begin{equation}
  \mbox{PSNR}(Y,\hat{Y}) = 10
  \log_{10}{\frac{\max_{\hat{Y}}^2}{\frac{1}{N}\sum_{i=0}^N (Y_i-\hat{Y}_i)^2
  }},
  \end{equation}
where $\max_{\hat{Y}}$ is the maximum possible value of the image intensities.
We also provide the Structural Similarity Index Measure (SSIM) of
\citet{Wang04SSIM}. It
ranges between -1 and 1, a larger score meaning a greater similarity between the
two images.

To measure the loss of sharpness between the true frame and the
prediction, we define the following sharpness measure based on the difference of gradients between two
images $Y$ and $\hat{Y}$:

\omitme{ratio of sums of gradients between two
images $Y$, $\hat{Y}$:
\begin{equation}
  \mbox{Sharp. ratio}(Y,\hat{Y}) = \frac{\sum_{i} \sum_{j} \nabla_iY + \nabla_jY  }
  {\sum_{i} \sum_{j}   \nabla_i\hat{Y} + \nabla_j\hat{Y}  }=
  \frac{\sum_{i} \sum_{j} |Y_{i,j}-Y_{i-1,j}| + |Y_{i,j}-Y_{i,j-1}|  }
  {\sum_{i} \sum_{j}  |\hat{Y}_{i,j}-\hat{Y}_{i-1,j}| + |\hat{Y}_{i,j}-\hat{Y}_{i,j-1}| }.
\end{equation}
A sharpness ratio of one indicates that the predicted image is as sharp as
the target, and lower scores indicate blurrier predictions. This measure has the
advantage of being easily interpretable, however it results in large score even
if the gradient of the prediction do not correspond to the gradient of the
target. We therefore introduce a second sharpness measure based on the
difference of the gradients:}
\omitme{ \begin{multline}
 \mbox{Sharpness}(Y,\hat{Y}) = \\ 10
  \log_{10}{\frac{\max_{\hat{Y}}^2}{\frac{1}{N}
\left( \sum_{i} \sum_{j}  ( |Y_{i,j}-Y_{i-1,j}| + |Y_{i,j}-Y_{i,j-1}| )- (\hat{Y}_{i,j}-\hat{Y}_{i-1,j}|+|\hat{Y}_{i,j}-\hat{Y}_{i,j-1}|)\right)}}.
\end{multline}}
\begin{equation}
 \mbox{Sharp. diff.}(Y,\hat{Y}) = 10
  \log_{10}{\frac{\max_{\hat{Y}}^2}{\frac{1}{N}
\left( \sum_{i} \sum_{j} | (  \nabla_iY + \nabla_jY )- (\nabla_i\hat{Y}+\nabla_j\hat{Y} )| \right)}}.
\end{equation}
where $\nabla_iY = |Y_{i,j}-Y_{i-1,j}|$ and $\nabla_jY = |Y_{i,j}-Y_{i,j-1}|$.

\begin{table}
    \caption{Comparison of the accuracy of the predictions on $10\%$ of the UCF101 test
      images. The different models have been trained given 4 frames to predict the next
       one. Similarity and sharpness measures evaluated only in the areas of movement.
    Our best model has been fine-tuned on UCF101 after the training on Sports1m.}
    \begin{center}
 \begin{tabular}{c|cc|c|cc|c|}
        \cline{2-7}
        & \multicolumn{3}{ |c| }{$1^{\text{st}}$ frame prediction scores} & \multicolumn{3}{ c| }{$2^{\text{nd}}$ frame prediction scores}  \\
 \cline{2-7}
 & \multicolumn{2}{ |c| }{Similarity} & Sharpness &
 \multicolumn{2}{ |c| }{Similarity} & Sharpness \\
 & PSNR & SSIM &  &  PSNR & SSIM   & \\
 \hline
% \multicolumn{1}{|c|}{ single sc. $\ell_2$ } & 26.9 & 0.85 & 24.7& 0.56 & 23.0
 % &0.82 & 24.3 & 0.49 \\
 \multicolumn{1}{|c|}{ single sc. $\ell_2$ } & 26.5 & 0.84  & 24.7   & 22.4& 0.82 & 24.2 \\
\multicolumn{1}{|c|}{ $\ell_2$ } & 27.6 & 0.86 & 24.7   & 22.5  &0.81 &24.2 \\
\multicolumn{1}{|c|}{ $\ell_1$  } & 28.7 & 0.88 & 24.8  & 23.8  & 0.83 &24.3 \\
\multicolumn{1}{|c|}{ GDL $\ell_1$} & 29.4  & 0.90 & 25.0 & 24.9  & 0.84 & 24.4
\\
\multicolumn{1}{|c|}{ GDL $\ell_1^*$} & 29.9  & 0.90 & 25.0 & 26.4 & 0.87 & 24.5 \\
\multicolumn{1}{|c|}{ Adv$^*$ } & 30.6 & 0.89 & 25.2 & 26.1 & 0.85 & 24.2  \\
\multicolumn{1}{|c|}{ Adv+GDL$^*$ } &  31.5 &  0.91 & {\bf 25.4} & 28.0 &  0.87 & {\bf 25.1} \\
\multicolumn{1}{|c|}{ Adv+GDL fine-tuned $^*$} & {\bf 32.0} &  0.92 &  {\bf 25.4} & {\bf 28.9}&  0.89 & 25.0 \\
%\multicolumn{1}{|c|}{ Adv+GDL fine-tuned $^*$} & {\bf 31.7} &  {\bf 0.93} &
%25.1 & {\bf 28.8}& {\bf 0.90} & 24.7 \\
\hline
\multicolumn{1}{|c|}{ Last input } & 28.6 & 0.89 & 24.6  & 26.3 &  0.87 &24.2  \\
\multicolumn{1}{|c|}{ Optical flow } & 31.6 & {\bf 0.93}  & 25.3  & 28.2  & {\bf 0.90} & 24.7 \\
\hline
 \end{tabular}\\
 $^*$ models fine-tuned on patches of size $64 \times 64$.
    \end{center}
    \label{tbl:compUCF101test}
\end{table}

\begin{figure}[htb]
  \caption{Our evaluation of the accuracy of future frames prediction only takes
    the moving areas of the images into account. Left: example of our frame
    predictions in a entire image with ground truth; Right: images masked with
    thresholded optical flow.}
  \begin{center}
    \includegraphics[width=0.12\textwidth]{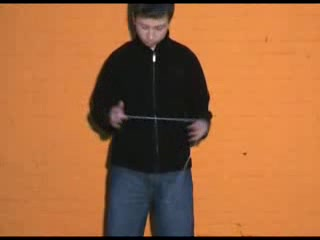}~
    \includegraphics[width=0.12\textwidth]{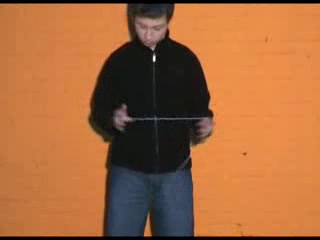}~
    \includegraphics[width=0.12\textwidth]{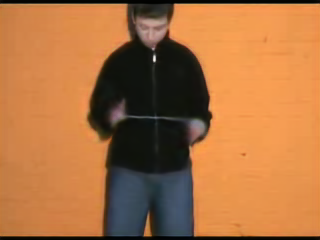}~
    \includegraphics[width=0.12\textwidth]{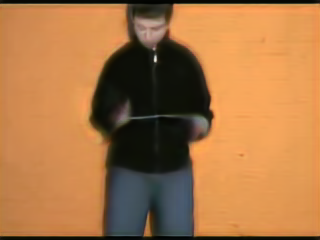}~
    \includegraphics[width=0.12\textwidth]{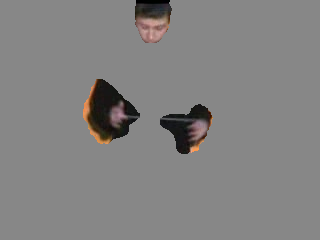}~
    \includegraphics[width=0.12\textwidth]{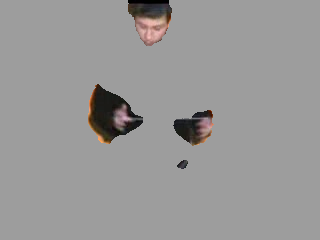}~
    \includegraphics[width=0.12\textwidth]{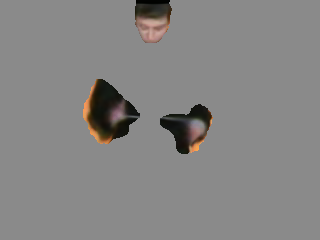}~
    \includegraphics[width=0.12\textwidth]{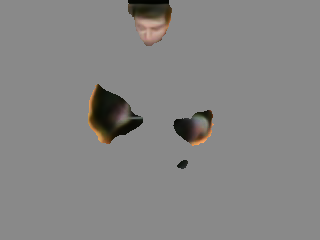}\\
\begin{tabular}{cccccccc}
    Target  &Target  &Adv+GDL& Adv+GDL& Masked & Masked & Masked & Masked \\
    image 1 & image 2 & Pred. 1 & Pred. 2&  Target 1 &Target 2 & Pred. 1 & Pred. 2\\
  \end{tabular}
    \end{center}
\label{flow_masks}
\end{figure}

As for the other measures, a larger score is better.  These quantitative
measures on 378 test videos from UCF101\footnote{We extracted from the test set
  list video files every 10 videos, starting at 1, 11, 21 etc.} are given in
Table \ref{tbl:compUCF101test}. As it is trivial to predict pixel values in
static areas, especially on the UCF101 dataset where most of the images are
still, we performed our evaluation in the moving areas as displayed in
Figure~\ref{flow_masks}.  To this end, we use the EpicFlow method of
\citet{Revaud2015Epic}, and compute the different quality measures only in the
areas where the optical flow is higher than a fixed threshold \footnote{We use
  default parameters for the Epic Flow computation, and transformed the .flo
  file to png using the Matlab code
  \url{http://vision.middlebury.edu/flow/code/flow-code-matlab.zip}. If at least
  one color channel is lower than 0.2 (image color range between 0 and 1), we
  replace the corresponding pixel intensity of the output and ground truth to 0,
  and compute similarity measures in the resulting masked images.}.
%These two separate evaluations allow us to
%compare the differents model on full image prediction, and to demonstrate the
%advantage of the proposed methods in comparison to a simple copy of the last
%input frame in the area of motion.
Similarity and sharpness measures computed on the whole images are given in Appendix.

The numbers clearly indicate that all strategies perform
better than the $\ell_2$ predictions in terms of PSNR, SSIM and sharpness. The
multi-scale model brings some improvement, but used with an $\ell_2$ norm, it
does not outperform simple frame copy in the moving areas. The $\ell_1$ model
improves the results, since it
replaces the mean by the median value of individual pixel
predictions. The GDL and adversarial predictions are
leading to further gains, and finally the combination of the multi-scale,
$\ell_1$ norm, GDL and adversarial training achieves the best PSNR,
SSIM and Sharpness difference measure.

It is interesting to note that while we showed that the $\ell_2$ norm was a poor metric
for training predictive models, the PSNR at test time is the worst for models
trained optimising the $\ell_2$ norm, although the PSNR is based on the $\ell_2$ metric.
We also include the baseline presented in
\citet{Ranzato2014videolanguage} -- courtesy of Piotr Dollar -- that extrapolates
the pixels of the next frame by propagating the optical flow from the previous ones.
%Finally, we fine-tuned our best model Adv+GDL on UCF101 to predict the second
%frame given the three input frames with the output of the first predicted frame.

%The best results are achieved by the
%adversarial model in term of Sharpness, and by the combined version of the
%adversarial training and GDL in term of PSNR. We note that, in this second
%setting, all generative feature maps are approximately 4 times larger.

\begin{figure}[htb]
  \caption{Results on 3 video clips from Sport1m. Training: 4 inputs, 1
    output. Second output
    computed recursively.}
  \begin{center}
%%x    \begin{tabular}{cccc}

%%       % input frames
%%       \multicolumn{2}{c}{\includegraphics[height=2.4cm]{inputs3.png}} \\
%%       \multicolumn{2}{c}{Input frames} \\
%%       % GT , L2
%%       \includegraphics[height=2.4cm]{Seq3_output2} & \includegraphics[height=2.4cm]{Seq3_2_L2} \\
%%       Ground truth & $\ell_2$ result \\
%%       % L1 , GDL
%%       \includegraphics[height=2.4cm]{Seq3_2_L1} & \includegraphics[height=2.4cm]{Seq3_2_TVL1} \\
%%       $\ell_1$ result & GDL $\ell_1$ result \\
%%       % adv, adv+gdl
%%       \includegraphics[height=2.4cm]{adv_3.png} & \includegraphics[height=2.4cm]{advgdl_3.png} \\
%%       Adversarial result & Adversarial+GDL result \\
%%     \end{tabular}\\

    \hspace*{-2.2cm}
    \begin{tabular}{c@{\ \ }c@{\ \ }c@{\ \ }c}
      % input frames , GT , L2
      \multicolumn{2}{l}{\includegraphics[height=1.65cm]{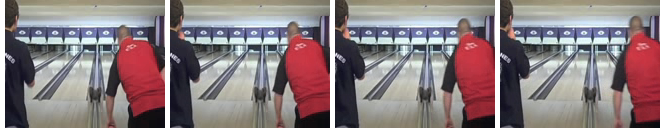}} & \includegraphics[height=1.65cm]{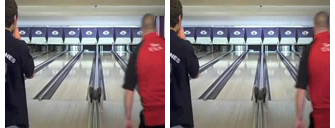} & \includegraphics[height=1.65cm]{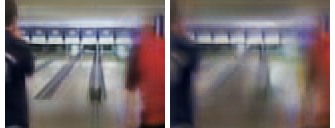} \\
      \multicolumn{2}{c}{Input frames} & Ground truth & $\ell_2$ result \\
      % L1 , GDL , adv, adv+gdl
      \includegraphics[height=1.65cm]{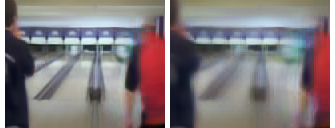} & \includegraphics[height=1.65cm]{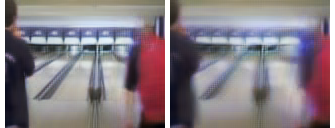} & \includegraphics[height=1.65cm]{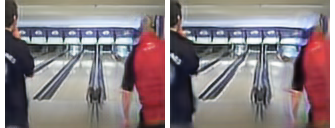} & \includegraphics[height=1.65cm]{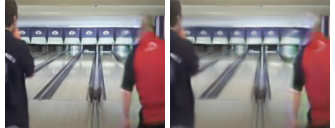} \\
      $\ell_1$ result & GDL $\ell_1$ result & Adversarial result & Adversarial+GDL result \\
    \end{tabular}\\

    \hspace*{-2.2cm}
    \begin{tabular}{c@{\ \ }c@{\ \ }c@{\ \ }c}
      % input frames , GT , L2
      \multicolumn{2}{l}{\includegraphics[height=1.65cm]{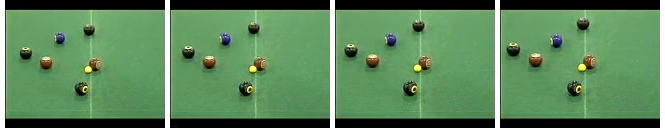}} & \includegraphics[height=1.65cm]{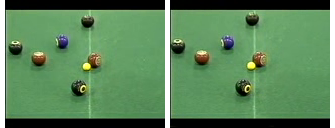} & \includegraphics[height=1.65cm]{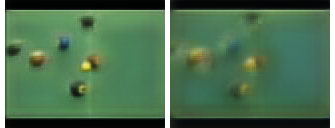} \\
      \multicolumn{2}{c}{Input frames} & Ground truth & $\ell_2$ result \\
      % L1 , GDL , adv, adv+gdl
      \includegraphics[height=1.65cm]{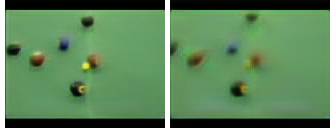} & \includegraphics[height=1.65cm]{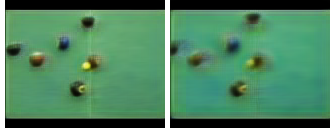} & \includegraphics[height=1.65cm]{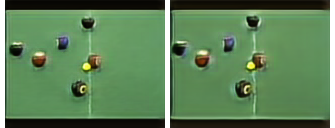} & \includegraphics[height=1.65cm]{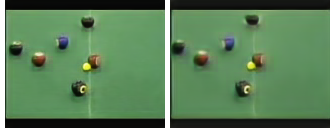} \\
      $\ell_1$ result & GDL $\ell_1$ result & Adversarial result & Adversarial+GDL result \\
    \end{tabular}\\

    \hspace*{-2.2cm}
    \begin{tabular}{c@{\ \ }c@{\ \ }c@{\ \ }c}
      % input frames , GT , L2
      \multicolumn{2}{l}{\includegraphics[height=1.65cm]{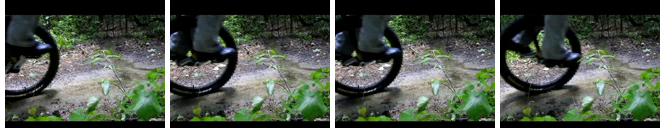}} & \includegraphics[height=1.65cm]{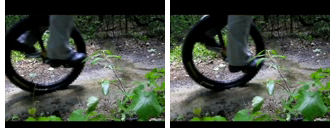} & \includegraphics[height=1.65cm]{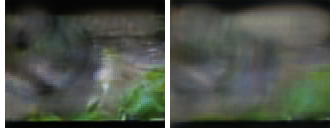} \\
      \multicolumn{2}{c}{Input frames} & Ground truth & $\ell_2$ result \\
      % L1 , GDL , adv, adv+gdl
      \includegraphics[height=1.65cm]{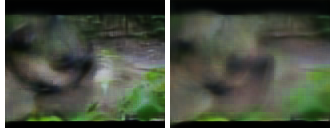} & \includegraphics[height=1.65cm]{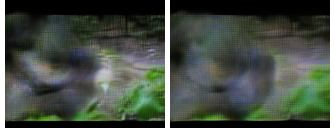} & \includegraphics[height=1.65cm]{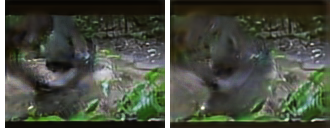} & \includegraphics[height=1.65cm]{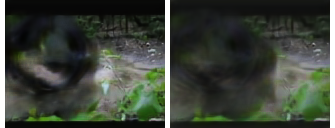} \\
      $\ell_1$ result & GDL $\ell_1$ result & Adversarial result & Adversarial+GDL result \\
    \end{tabular}\\

    \end{center}
\label{fig:Seq3}
\end{figure}

Figure \ref{fig:Seq3} shows results on test sequences from the Sport1m
dataset, as movements are more visible in this dataset.

%**************************************
\subsection{Comparison to \citet{Ranzato2014videolanguage}}
%**************************************

\begin{figure}[htb]
  \caption{Comparison of results on the Basketball Dunk and Ice Dancing
      clips from UCF101 appearing in \citep{Ranzato2014videolanguage}. We
      display 2 frame predictions for each method along with 2 zooms of each
      image. The PSNR and SSIM values are computed in the moving areas of the
      images (More than the 2/3 of the pixels in these examples). The values in parenthesis correspond to the second frame
      predictions measures.}
  \begin{center}
\hspace*{-0.5cm} \begin{tabular}{cc}
  \includegraphics[width=0.25\textwidth]{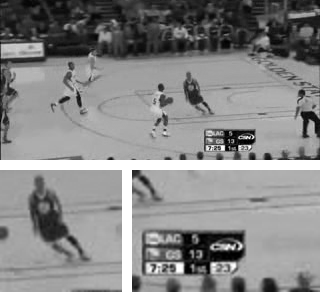}
  \includegraphics[width=0.25\textwidth]{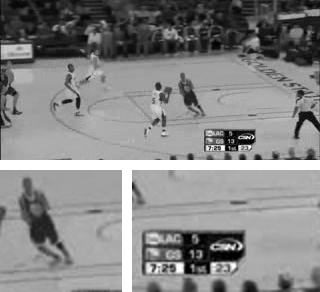}&
  \includegraphics[width=0.25\textwidth]{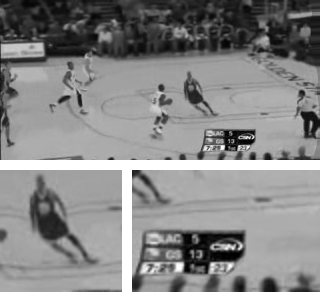}
  \includegraphics[width=0.25\textwidth]{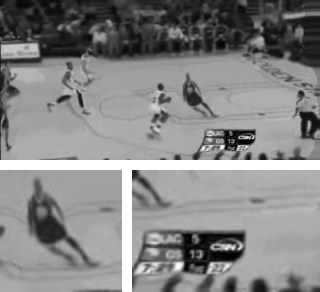}\\
  Target & Prediction using a constant optical flow \\
 & PSNR = 25.4 (18.9), SSIM = 0.88 (0.56) \\
  \end{tabular}\\
\hspace*{-0.5cm} \begin{tabular}{cc}
  \includegraphics[width=0.25\textwidth]{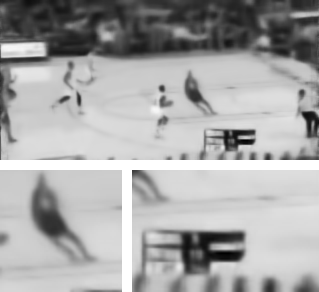}
  \includegraphics[width=0.25\textwidth]{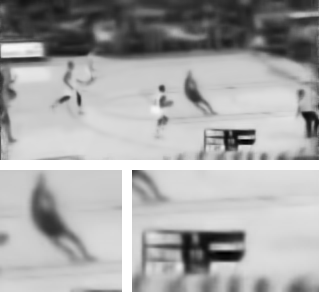} &
  \includegraphics[width=0.25\textwidth]{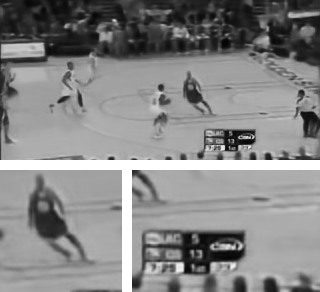}
  \includegraphics[width=0.25\textwidth]{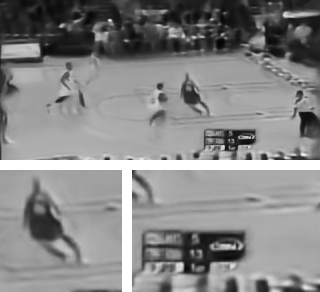}\\
  Ranzato et al. result &  Adv GDL $\ell_1$ result\\
 PSNR = 16.3 (15.1), SSIM = 0.70 (0.55)& PSNR = 26.7 (19.0), SSIM = 0.89 (0.59)\\

\end{tabular}
  \end{center}
~\\
  \begin{center}
\hspace*{-0.5cm} \begin{tabular}{cc}
  \includegraphics[width=0.25\textwidth]{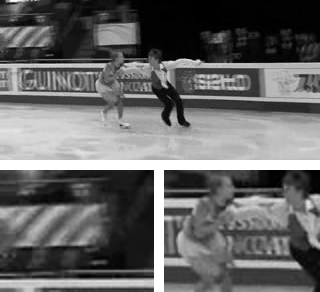}
  \includegraphics[width=0.25\textwidth]{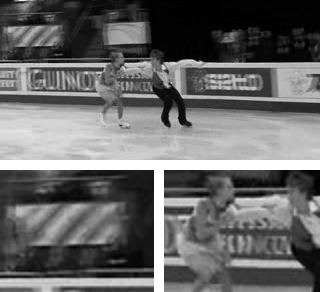}&
  \includegraphics[width=0.25\textwidth]{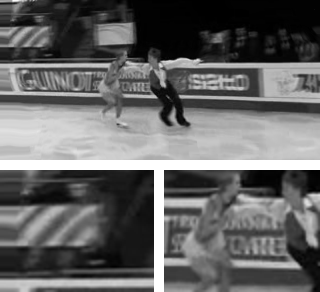}
  \includegraphics[width=0.25\textwidth]{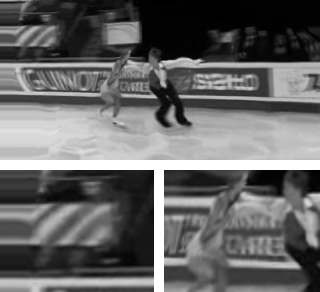}\\
  Target & Prediction using a constant optical flow \\
& PSNR = 24.7 (20.6), SSIM = 0.84 (0.72)\\
\end{tabular}\\
\hspace*{-0.5cm} \begin{tabular}{cc}
  \includegraphics[width=0.25\textwidth]{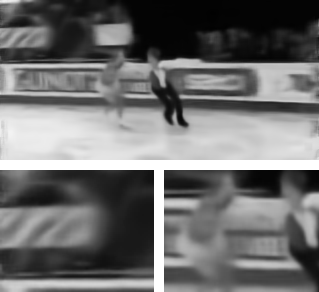}
  \includegraphics[width=0.25\textwidth]{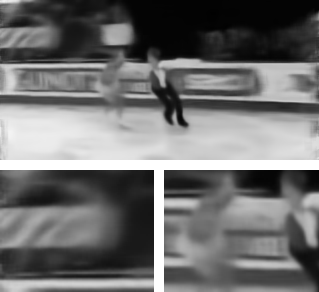} &
  \includegraphics[width=0.25\textwidth]{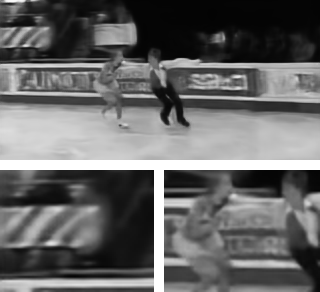}
  \includegraphics[width=0.25\textwidth]{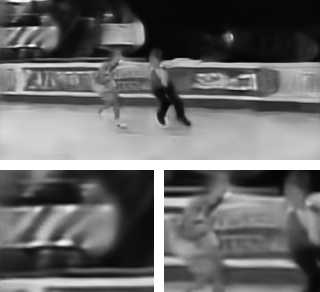}\\
Ranzato et al. result &  Adv GDL $\ell_1$ result\\
PSNR = 20.1 (17.8), SSIM = 0.72 (0.65) & PSNR = 24.6 (20.5), SSIM = 0.81 (0.69)\\
\end{tabular}
\end{center}
\label{comparisonMarcAurelio}
\end{figure}

In this section, we compare our results to \citep{Ranzato2014videolanguage}.
To obtain grayscale images, we make RGB predictions and extract the Y channel
of our Adv+GDL model.
% \omitme{and selected the
%two publicly UCF101 videos presenting their results that contain the largest
%amount of movements.}
\citet{Ranzato2014videolanguage} images are generated by
averaging 64 results obtained using different tiling to avoid
a blockiness effect, however creating instead a blurriness effect. We compare
the PSNR and SSIM values on the first predicted images of Figure~\ref{comparisonMarcAurelio}.

\hspace*{-0.1cm}

We note that the results of Ranzato et al. appear slightly lighter than our
results because of a normalization that does not take place in the
original images, therefore the errors given here are not reflecting the full
capacity of their approach.  We tried to apply the blind deconvolution method of
\citet{Krishnan2011blind} to improve Ranzato et al. and our different results. As
expected, the obtained sharpness scores are higher, but the image similarity
measures are deteriorated because often the contours of the predictions do not
match exactly the targets.
More importantly, Ranzato et al. results appear to be more static in moving areas.
Visually, the optical flow result appears similar to the target, but a closer look at
thin details reveals that lines, heads of people are bent or squeezed.

\section{Conclusion}
%**********************************************************************************

We provided a benchmark of several strategies for next frame prediction, by
evaluating the quality of the prediction in terms of Peak Signal to Noise Ratio,
Structural Similarity Index Measure
and image sharpness. We display our results on small UCF video clips at
\url{http://cs.nyu.edu/~mathieu/iclr2016.html}. The presented architectures and
losses may be used as building blocks for more sophisticated prediction models,
involving memory and recurrence. Unlike most optical flow algorithms, the model
is fully differentiable, so it can be fine-tuned for another task if necessary.
Future work will deal with the evaluation of
the classification performances of the learned representations in a weakly
supervised context, for instance on the UCF101 dataset. Another extension of
this work could be the combination of the current system with optical flow
predictions. Alternatively, we could replace optical flow
predictions in applications that does not explicitly require optical flow but
rather next frame predictions. A simple example is causal (where the next frame
is unknown) segmentation of video streams.

%For instance, the work of \citet{Miksik2012} smoothes
%the successive 2D frame segmentations temporally by relying on optical flow
%predictions.

\omitme{
Future work: RGB loss function taking the max error out of the three color
channels, or change of color space (i.e. YUV or LAB)
}

\subsubsection*{Acknowledgments}
We thank Florent Perronnin for fruitful
discussions, and Nitish Srivastava, Marc'Aurelio Ranzato and Piotr Doll\'ar for
providing us their results on some video sequences.

% A remettre a la fin Du Tran, and Marc'Aurelio Ranzato

\bibliography{Mathieu2016ICLRNextFramePred}
\bibliographystyle{iclr2016_conference}

\newpage

\section{Appendix}

%**************************************
\subsection{Predicting the eight next frames }
%**************************************

In this section, we trained our different multi-scale models -- architecture
described in Table \ref{tbl:network2}--  with $8$ input frames to predict
$8$ frames simultaneously.
Image similarity measures are given between the ground truth and the predictions
in Table~\ref{8outputs}.

\begin{table}[htb]
  \caption{Network architecture}
  \begin{center}
    {\bf Models 8 frames in input -- 8 frames in output}\\
    \begin{tabular}{ccccc}
      \hline
      Generative network scales  & $G_1$ & $G_2$&  $G_3$ & $G_4$ \\
      \hline
      Number of feature maps &  16, 32, 64 & 16, 32, 64 & 32, 64, 128 & 32, 64, 128, 128 \\
      Conv. kernel size & 3, 3, 3, 3 & 5, 3, 3, 3 & 5, 5, 5, 5 & 7, 5, 5, 5, 5 \\
      \hline
      Adversarial network scales  & $D_1$ & $D_2$&  $D_3$ & $D_4$ \\
      \hline
      Number of feature maps &  16 & 16, 32, 32 & 32, 64, 64 & 32, 64, 128,
      128 \\
      Conv. kernel size (no padding) & 3 & 3, 3, 3 & 5, 5, 5 & 7, 7, 5, 5 \\
      Fully connected & 128, 64 & 256, 128 & 256, 128 & 256, 128 \\
      \hline
    \end{tabular}\\
    \end{center}
  \label{tbl:network2}
\end{table}

For the first and eighth predicted frames,
the numbers clearly indicate that all strategies perform
better than the $\ell_2$ predictions in terms of PSNR and sharpness. The $\ell_1$ model, by
replacing the mean intensity by the median value in individual pixel
predictions, allows us to improve results. The adversarial predictions are
leading to further gains, and finaly the GDL allows
the predictions to achieve the best PNSR and sharpness.
We note that the size of the network employed in the
simultaneous prediction configuration is smaller than in the unique frame
prediction setting.

\begin{table}[h]
    \caption{Comparison of the accuracy of the predictions on $10\%$ of the UCF101 test
      images. The different models have been trained given 8 frames to predict
      the 8 next ones.}
    \begin{center}
\begin{tabular}{c|cc|c|cc|c|}
        \cline{2-7}
        & \multicolumn{3}{ |c| }{$1^{\text{st}}$ frame prediction scores} & \multicolumn{3}{ c| }{$8^{\text{th}}$ frame prediction scores}  \\
 \cline{2-7}
 & \multicolumn{2}{ |c| }{Similarity} & Sharpness & \multicolumn{2}{ |c| }{Similarity} & Sharpness \\
 & PSNR & SSIM &  &  PSNR & SSIM &   \\
 \hline
%\multicolumn{1}{|c|}{ L2  4000 ep.} & 15.7 & 0.54 & 13.9 & 0.43\\
%\multicolumn{1}{|c|}{ L2  3000 ep.} & 16.5 & 0.52 & 15.2 & 0.41\\
\multicolumn{1}{|c|}{ $\ell_2$ } & 18.3 & 0.59 & 17.5  & 15.4 & 0.51 & 17.4  \\ % 2000 ep.
\multicolumn{1}{|c|}{ Adv } & 21.1 & 0.61 & 17.6  & 17.1 & 0.52 & 17.4  \\ % 4000 ep.
\multicolumn{1}{|c|}{ $\ell_1$  } & 21.3 & 0.66 & 17.7  & 17.0 & 0.55 & 17.5  \\ % 3000 ep.
%\multicolumn{1}{|c|}{ GDL L2 1000 ep.} &
%\multicolumn{1}{|c|}{ GDL L1 big 2000 ep.} & 18.7 & 0.92 & 17.8 & 0.83  \\
%\multicolumn{1}{|c|}{ GDL L1 2000 ep.} & 21.4 & 0.96 & 19.8 & 0.87  \\
 \multicolumn{1}{|c|}{ GDL $\ell_1$} & 21.4 & 0.69 & 17.9  & 17.7 & 0.58 & 17.5 \\ % 2000 ep
%% to find \multicolumn{1}{|c|}{ GDL $\ell_1$} & {\bf 21.8} & {\bf 0.87} & {\bf 19.2} & {\bf 0.79}  \\% 3000 ep.
%\multicolumn{1}{|c|}{ GDL $\ell_1$ large} & 21.6 & {\bf 0.88} & {\bf
%  20.7} & 0.77  \\ % 1000 ep.
\multicolumn{1}{|c|}{Last input} & 30.6 & 0.90 & 22.3  & 21.0 & 0.74 & 18.5 \\
\hline
\end{tabular}\\
    \end{center}
    \label{8outputs}
\end{table}

\begin{figure}[htb]
  \caption{Results on a UCF101 video using a large GDL-$\ell_1$ model. Training: 8
    inputs, 8 outputs. First line: target, second line: our predictions.}
  \begin{center}
    \includegraphics[width=1\textwidth]{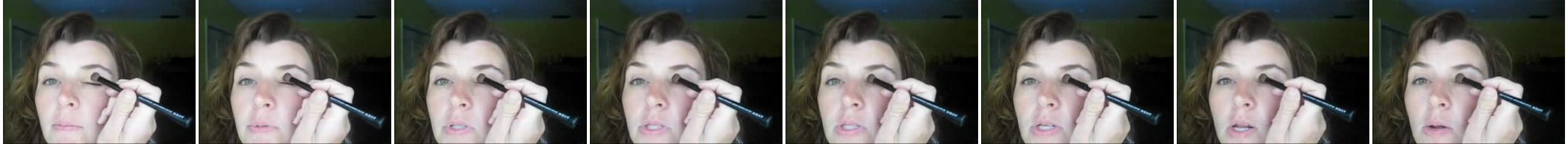} \\
    ~ \\
    \includegraphics[width=1\textwidth]{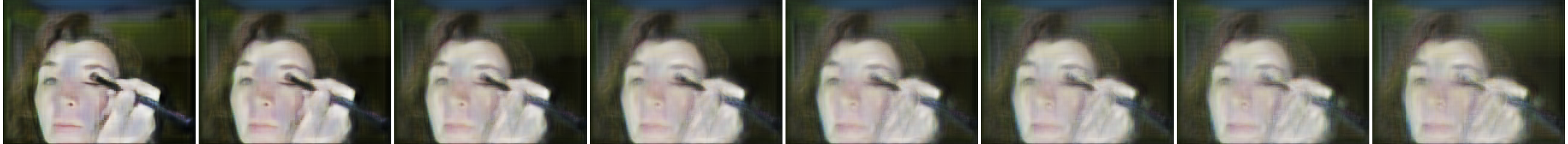} \\
    \end{center}
\label{fig:8outputs}
\end{figure}

Figure~\ref{fig:8outputs} shows a generation result of eight frames simultaneously,
using a large version of the GDL $\ell_1$ model in which all the number of
feature maps were multiplied by four.

Compared to the recursive frame prediction as employed in the rest of the paper,
predicting several input simultaneouly leads to better long term results but
worst shorter term ones.  The gap between the two performances could be reduced
by the design of time multi-scale strategies.

%**************************************
\subsection{Comparison to the LSTM approach of \citet{SrivastavaMS15} }
%**************************************

\begin{figure}[htb]
\caption{Comparison of different methods to predict $32\times 32$ patches from UCF101.
  The LSTM lines are from~\citep{SrivastavaMS15}. Baseline results with pure $\ell_1$
  and $\ell_2$ losses are shown in Appendix.}
\vspace*{-0.2cm}
  \begin{center}
\hspace*{-0.9cm} \begin{tabular}{ccc}
     % Sequence 1 & &  Sequence 2 \\
\includegraphics[width=0.43\textwidth]{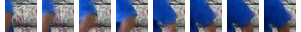} & Input &\includegraphics[width=0.43\textwidth]{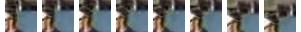} \\
\includegraphics[width=0.43\textwidth]{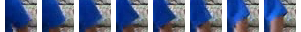} & Ground truth & \includegraphics[width=0.43\textwidth]{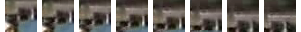}\\
\includegraphics[width=0.43\textwidth]{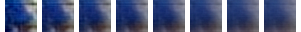} & LSTM 2048& \includegraphics[width=0.43\textwidth]{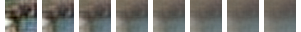}\\
\includegraphics[width=0.43\textwidth]{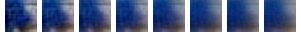} & LSTM 4096& \includegraphics[width=0.43\textwidth]{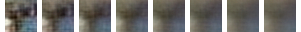}\\
\includegraphics[width=0.43\textwidth]{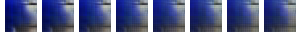} & GDL $\ell_1$ & \includegraphics[width=0.43\textwidth]{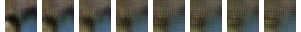}\\
\includegraphics[width=0.43\textwidth]{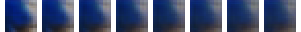} & $\ell_1$ & \includegraphics[width=0.43\textwidth]{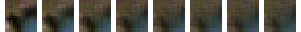}\\
\includegraphics[width=0.43\textwidth]{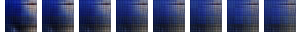} & GDL $\ell_2$ & \includegraphics[width=0.43\textwidth]{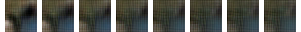}\\
\includegraphics[width=0.43\textwidth]{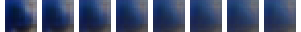} &  $\ell_2$ &  \includegraphics[width=0.43\textwidth]{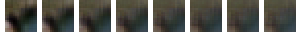}\\
\includegraphics[width=0.43\textwidth]{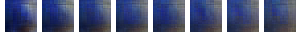} & Adversarial   & \includegraphics[width=0.43\textwidth]{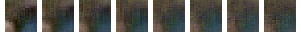}\\
\includegraphics[width=0.43\textwidth]{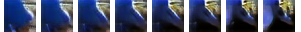} & Adv. recursive  &
\includegraphics[width=0.43\textwidth]{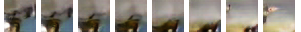}\\
\includegraphics[width=0.43\textwidth]{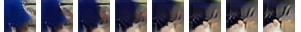} & Adv. rec. + GDL  &
\includegraphics[width=0.43\textwidth]{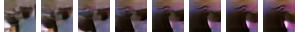}\\
\includegraphics[width=0.43\textwidth]{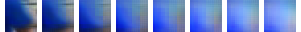} & GDL $\ell_1$ recursive  & \includegraphics[width=0.43\textwidth]{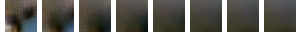}\\
\includegraphics[width=0.43\textwidth]{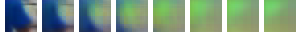} &  $\ell_1$ rec. & \includegraphics[width=0.43\textwidth]{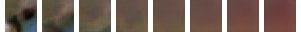}\\
\includegraphics[width=0.43\textwidth]{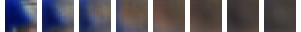} & $\ell_2$ rec.  & \includegraphics[width=0.43\textwidth]{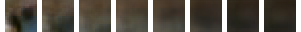}\\
\end{tabular}\\
\end{center}
\label{fig:comp_LSTM}
\end{figure}

Figure~\ref{fig:comp_LSTM} shows a comparison with predictions based on LSTMs
using  sequences of patches from \citep{SrivastavaMS15}.
The model ranking established on UCF101 in terms of sharpness and PSNR remains
unchanged on the two sequences. %The actual numbers are shown in Table~\ref{tbl:compLSTM}
%The model ranking established on UCF101 in terms of sharpness and PSNR remains
%unchanged on the two sequences.
When we employ the setting 8 inputs-8 output described in Table \ref{tbl:network2},
 %the LSTM approach ranks second behind the GDL-$\ell_1$ model in term of PSNR.
we note that the LSTM first frame prediction is sharper than our models
predictions, however when looking at a longer term future, our gradient
difference loss leads to sharper results.
Comparing visually the GDL $\ell_1$ and GDL $\ell_2$, we notice that the
predictions suffer from a chessboard effect in the $\ell_2$ case.
%probably because the image gradient of the prediction tends to be equal to
%the average of the image gradient, which is larger than the median gradient.
% TODO: i don't understand this sentence. I removed it.
On the other hand, when employing the recursive strategy (4 inputs, 1 output),
the adversarial training lead to much sharper predictions. It does not look
like anything close to the ground truth on the long term, but it remains
realistic.

\subsection{Additional results on the UCF101 dataset}

We trained the model described in Table \ref{tbl:network} with our different losses to
predict 1 frame from the 4 previous ones.
We provide in Table \ref{tab:full} similarity (PSNR and SSIM) and sharpness
measures between the different tested models predictions and frame to
predict. The evaluation is performed on the full images but is not really
meaningful because predicting the future location of static pixels is most
accurately done by copying the last input frame.

\begin{table}[h]
  \caption{Comparison of the accuracy of the predictions on $10\%$ of the UCF101 test
      images. The different models have been trained given 4 frames to predict the next
      one. Similarity and sharpness measures on full images.}
  \begin{center}
       \begin{tabular}{c|cc|c|cc|c|}
        \cline{2-7}
        & \multicolumn{3}{ |c| }{$1^{\text{st}}$ frame prediction scores} & \multicolumn{3}{ c| }{$2^{\text{nd}}$ frame prediction scores}  \\
 \cline{2-7}
 & \multicolumn{2}{ |c| }{Similarity} & Sharpness & \multicolumn{2}{ |c| }{Similarity} & Sharpness \\
 & PSNR & SSIM & &  PSNR & SSIM &   \\
 \hline
 %\multicolumn{1}{|c|}{ single sc. $\ell_2$ } & 19.6 & 0.61 & 17.8 & 0.35 & 14.9
 %& 0.51 & 17.4 & 0.24 \\
 \multicolumn{1}{|c|}{ single sc. $\ell_2$ } & 19.0 & 0.59  & 17.8   & 14.2 & 0.48 &17.5     \\
\multicolumn{1}{|c|}{ $\ell_2$ } & 20.1 & 0.64 & 17.8  & 14.1 & 0.50 & 17.4 \\
\multicolumn{1}{|c|}{ $\ell_1$  } & 22.3 & 0.74 & 18.5  & 16.0 & 0.56 & 17.6  \\
\multicolumn{1}{|c|}{ GDL $\ell_1$} & 23.9 & 0.80 & 18.7  & 18.6 & 0.64 & 17.7 \\
\multicolumn{1}{|c|}{ Adv } & 24.4 & 0.77 & 18.7  & 18.9 & 0.59& 17.3 \\
\multicolumn{1}{|c|}{ Adv+GDL } &  27.2 & 0.83 &  19.6 & 22.6&  0.72 &  18.5 \\
\multicolumn{1}{|c|}{ Adv+GDL fine-tuned} & 29.6 & {\bf 0.90} & 20.3 &{\bf 26.0} & 0.83  & 19.4  \\
\hline
\multicolumn{1}{|c|}{ Last input } & {\bf 30.0} & {\bf 0.90} & {\bf 22.1}  &25.8 & {\bf 0.84} & {\bf 20.3}  \\
\hline
%\multicolumn{1}{|c|}{ Difference GDL } & {\bf 30.5} & {\bf 0.90} & {\bf 22.3}  &  &   \\
\hline
       \end{tabular}\\
       \end{center}
\label{tab:full}
\end{table}

\end{document}